*Editorial*

# Editorial for the First Workshop on Mining Scientific Papers: Computational Linguistics and Bibliometrics

Iana Atanassova, Marc Bertin, Philipp Mayr

**Introduction**

The open access movement in scientific publishing and search engines like Google Scholar has made scientific articles more broadly accessible. During the last decade, the availability of scientific papers in full text has become more and more widespread thanks to the growing number of publications on online platforms such as ArXiv and CiteSeer [1]. The efforts to provide articles in machine-readable formats and the rise of Open Access publishing have resulted in a number of standardized formats for scientific papers (such as NLM-JATS, TEI, DocBook), full text datasets for research experiments (PubMed, JSTOR, etc.) and corpora (iSearch, etc.). At the same time, research in the field of Natural Language Processing have provided a number of open source tools for versatile text processing (e.g. NLTK, Mallet, OpenNLP, CoreNLP, Gate [2], CiteSpace [3]).

Scientific papers are highly structured texts and display specific properties related to their references but also argumentative and rhetorical structure. Recent research in this field has concentrated on the construction of ontologies for citations and scientific articles (e.g. CiTO [4], Linked Science [5]) and studies of the distribution of references [6]. However, up to now full text mining efforts are rarely used to provide data for bibliometric analyses. While Bibliometrics traditionally relies on the analysis of metadata of scientific papers (see e.g. a recent special issue on Combining Bibliometrics and Information Retrieval edited by Mayr & Scharnhorst, 2015 [7]), we will explore the ways full-text processing of scientific papers and linguistic analyses can play. With this workshop we like to discuss novel approaches and provide insights into scientific writing that can bring new perspectives to understand both the nature of citations and the nature of scientific articles. The possibility to enrich metadata by the full text processing of papers offers new fields of application to Bibliometrics studies.

Working with full text allows us to go beyond metadata used in Bibliometrics. Full text offers a new field of investigation, where the major problems arise around the organization and structure of text, the extraction of information and its representation on the level of metadata. Furthermore, the study of contexts around in-text citations offers new perspectives related to the semantic dimension of citations. The analyses of citation contexts and the semantic categorization of publications will allow us to rethink co-citation networks, bibliographic coupling and other bibliometric techniques.

**Workshop outline**

The workshop "Mining Scientific Papers: Computational Linguistics and Bibliometrics" (CLBib 2015)[1], co-located with the 15th International Society of Scientometrics and Informetrics Conference (ISSI 2015)[2], brought together researchers in Bibliometrics and Computational Linguistics in order to study the ways Bibliometrics can benefit from large-scale text analytics and sense mining of scientific papers, thus exploring the interdisciplinarity of Bibliometrics and Natural Language Processing (NLP). The goals of the workshop were to answer questions like: How can we enhance author network analysis and Bibliometrics using data obtained by text analytics? What insights can NLP provide on the structure of scientific writing, on citation networks, and on in-text citation analysis?

---

[1] http://www.gesis.org/en/events/conferences/issi-workshop-2015/
[2] http://issi2015.org/

The workshop topics included the following:
- Linguistic modelling and discourse analysis for scientific texts
- User interfaces, text representations and visualizations
- Structure of scientific articles (discourse / argumentative / rhetorical / social)
- Scientific corpora and paper standards
- Act of citations, in-text citations and Content Citation Analysis
- Co-citation and bibliographic coupling
- Text enhanced bibliographic coupling
- Terminology extraction
- Text mining and information extraction
- Scholarly information retrieval
- Ontological descriptions of scientific content
- Knowledge extraction

The call for papers attracted many contributions, showing a large interest in these topics in the community. After a selection process six papers were presented at the workshop. All papers have been peer-reviewed by at least two reviewers from the Program Committee. The following section briefly outlines the papers that were presented.

**Overview of the papers**

Natural language processing methods can be applied to scientific corpora in various domains in order to obtain data and gain insights on the domain and synthesize the information present in the documents. In the paper "NLP4NLP: Applying NLP to scientific corpora about written and spoken language processing" by **Gil Francopoulo, Joseph Mariani** and **Patrick Paroubek** [8], the authors report the creation of a large corpus of papers in Natural Language Processing and the processing of the corpus using methods and tools in the same field. The creation of large-scale scientific corpora that are accessible in interoperable formats, such as RDF, is an important step towards the development of NLP tools dedicated to scientific writing.

From the point of view of information extraction and text mining, the paper "Accurate Keyphrase Extraction from Scientific Papers by Mining Linguistic Information" by **Mounia Haddoud, Aïcha Mokhtari, Thierry Lecroq** and **Saïd Abdeddaïm** [9], proposes a method for keyphrase extraction from scientific papers using a supervised machine learning algorithm with linguistically motivated features, and more specifically a noun-phrase filter. The authors experiment with the SemEval-2010/Task-5 dataset and report that the use of linguistically motivated features in the classifier improves the system's ability to extract correct keyphrases compared to other similar systems.

**Bilal Hayat, Muhammad Rafi, Arsal Jamal, Raja Sami Ur Rehman, Muhammad Bilal Alam** and **Syed Muhammad Zubair Alam** propose to classify citations as sentiment positive or sentiment negative in the paper "Classification of Research Citations (CRC)" [10]. They use a sentiment lexicon with Naïve-Bayes Classifiers for sentiment analysis. Their algorithm is evaluated on a manually annotated and class labelled collection of 150 research papers from the domain of computer science and preliminary results show an accuracy of 80%. They plan to provide a web portal to assist the scholars in the automatic searching and downloading of citing papers, for a seed paper, and the classification of citing papers into sentiment categories.

In their paper "Using noun phrases extraction for the improvement of hybrid clustering with text- and citation-based components. The example of 'Information System Research'", **Bart Thijs, Wolfgang Glänzel** and **Martin Meyer** [11] propose a method for document clustering by the extraction of noun phrases from abstracts and titles to improve the measurement of the lexical component, and using bibliographic coupling for the citation component. They show in the hybrid clustering approach, removing all single term shingles provides the best results at the level of computational feasibility, comparability with bibliographic coupling and also in a community detection application.

The problem of automatic terminology extraction from scientific corpora is addressed in the paper "The Termolator: Terminology Recognition based on Chunking, Statistical and Search-based Scores" by **Adam Meyers, Yifan He, Zachary Glass** and **Olga Babko-Malaya** [12]. They propose the Termolator system, which combines several different approaches to extract chunks from texts. The authors examine several metrics for the ranking of the extracted terms and analyse the effects of these metrics on a corpus of 5000 patents on a specific topic. They report an accuracy of about 86% among the top 5000 terms.

The paper "Towards Authorship Attribution for Bibliometrics using Stylometric Features" by **Andi Rexha, Stefan Klampfl, Mark Kröll** and **Roman Kern** [13] describes a new text segmentation algorithm to identify potential author changes within the main text of a scientific article. Their approach captures stylistic changes in papers and adopts different stylometric features like type-token-ratio, hapax-legomena or average-word-length. A preliminary evaluation of a small subset of PubMed shows that the more authors an article has the more potential author changes are identified. In the paper the authors illustrate the style change among papers visually.

**Outlook**

This first workshop raises some important questions around the interactions between Bibliometrics and NLP. From an application point of view, the adequate tools need to be developed for the processing, information extraction and text mining of scientific corpora in order to produce new metadata or populate ontologies in the context of Semantic web. Tools for the efficient processing and conversion of PDF files are also necessary, allowing to produce corpora of scientific papers as structured documents, for example using XML schemas like JATS[3] that can be directly analysed by natural language processing modules. The main applications, beyond Bibliometrics, are in the field of Semantic Web.

The large-scale text analytics on scientific papers can also have an impact on the theory of citation, and contribute to define the nature of scientific citations. Text enhanced bibliometric approaches can provide ground for the development of new linguistics models related to the phenomenon of citation, involving new breakthroughs in this field. This workshop is the first step to foster the reflection on the interdisciplinarity and the benefits that the two disciplines Bibliometrics and Natural Language Processing can drive from it.

---

[3] http://jats.nlm.nih.gov/